\documentclass{article}

\usepackage{arxiv}

\usepackage[utf8]{inputenc} % allow utf-8 input
\usepackage[T1]{fontenc}    % use 8-bit T1 fonts
\usepackage{hyperref}       % hyperlinks
\usepackage{url}            % simple URL typesetting
\usepackage{booktabs}       % professional-quality tables
\usepackage{amsfonts}       % blackboard math symbols
\usepackage{amsmath}
\usepackage{nicefrac}       % compact symbols for 1/2, etc.
\usepackage{microtype}      % microtypography
\usepackage{lipsum}
\usepackage{graphicx}
\usepackage{csvsimple}

\graphicspath{ {./figures/} }

\DeclareMathAlphabet\mathbfcal{OMS}{cmsy}{b}{n}
\DeclareMathAlphabet{\altmathcal}{OMS}{cmsy}{m}{n}
\newcommand{\x}{x}
\newcommand{\y}{y}
\newcommand{\eMeters}{y_{m}}
\newcommand{\eTryTackle}{y_{tt}}
\newcommand{\eTrySet}{y_{ts}}
\newcommand{\winProb}{y_{w}}
\newcommand{\scoreline}{y_{s}}
\newcommand{\play}{y_{p}}
\newcommand{\inputVec}{\mathbf{x}}
\newcommand{\parameters}{\mathbf{t}}

\title{Rugby-Bot: Utilizing Multi-Task Learning \& Fine-Grained Features for Rugby League Analysis}

\author{
  Matthew Holbrook\\
  STATS, LLC\\
  Chicago, IL\\
  \texttt{mholbrook@stats.com} \\
   \And
  Jennifer Hobbs\\
  STATS, LLC\\
  Chicago, IL\\
  \texttt{jhobbs@stats.com} \\
  \And
  Patrick Lucey\\
  STATS, LLC\\
  Chicago, IL\\
  \texttt{plucey@stats.com} \\
}

\begin{document}
\maketitle
\begin{abstract}
Sporting events are extremely complex and require a multitude of metrics to accurate describe the event.
When making multiple predictions, one should make them from a single source to keep consistency across the predictions. 
We present a multi-task learning method of generating multiple predictions for analysis via a single prediction source.
To enable this approach, we utilize a fine-grain representation using fine-grain spatial data using a wide-and-deep learning approach.
Additionally, our approach can predict distributions rather than single point values. We highlighted the utility of our approach on the sport of Rugby League and call our prediction engine “Rugby-Bot”. 
\end{abstract}

% keywords can be removed

\section{Introduction}
\label{sec:intro}
For the past decade in sports analytics, the holy grail has been to find “the one best metric” which can best capture the performance of players and teams through the lens of winning. For example, expected metrics such as wins-above replacement (WAR) in baseball~\cite{whatisWAR}, expected point value (EPV) and efficiency metrics in basketball~\cite{Cervone2014POINTWISEP, oliver_BasketballOnPaper} and expected goal value in soccer~\cite{lucey2014quality} are used as the gold-standard in team and player analysis. In American Football, Defense-Adjusted Value Over Average (DVOA)~\cite{dvoa}, is perhaps the most respected and utilized advanced metric in the NFL which utilizes both an expected value and efficiency metric, and analyzes the value of a play compared to expected for every play and also normalizes for team, drive and game context. Win probability has also been widely used as the go-to metric for analyzing plays and performance across sports~\cite{inpredictableWP, iWinRNFL, statsWP}.

Although the works mentioned above are all impressive and have improved analysis and decision making across sports, we believe the overall hypothesis of having a single metric to explain all performance is limiting. By its very nature, sport is extremely complex and lends itself to many queries for different contexts and temporal resolutions. Instead of having a single predictor (or metric), we believe we need many predictors to enable these analyses.  For example, if a team has won the title, we want predictors that highlight their dominant attributes across the season (i.e., offensively and/or defensively strong, and if so which locations and times). Or if a team has won or lost a single match, we want predictors which highlight which plays or play was decisive or of note. 

While we believe that a multitude of predictors are required to enable multi-resolution analyses, we believe there should be a single source or brain that generates these predictors. For example, if we are predicting the expected point value of a drive in American Football and that jumps based on a big play, that should be correlated/co-occurring with a jump with the win-probability – otherwise this would possibly be contradictory and cause the user not to trust the model. Additionally, we believe the predictors should go beyond “expected metrics” which compare solely to the league average and produce distributions to enable deeper analysis (i.e., instead of comparing to the league average, why can’t we compare against the top 20\% or 10\% of teams). To effectively do this we need four ingredients:  i) a fine-grained representation of performance at the play-level which can be aggregated up to different resolutions, ii) fine-grain spatial data to enable the representation, iii) a multi-task learning approach where we train the predictors simultaneously, and iv) a method which can generate distributions instead of just point values.  

In this paper, we show how we can do this. We focus on the sport of Rugby League which has an abundance of spatial information available across many seasons. Specifically, we focus our analysis on the 2018 NRL Season which was the closest regular season in the 110-year history of Rugby League in Australia (and the history of all sport). Our single source prediction engine – which we call Rugby-Bot – showcases how our engine can highlight how we can uncover how the league was won.  In Figure 1 we showcase Rugby-Bot and in the next section we describe it.

\section{What is Rugby League and What is Rugby-Bot?}
Rugby League is a continuous and dynamic game played between 2 teams of 13 players (7 forwards and 6 backs) across 2x40 minute halves with the goal of “scoring a try” and obtaining points by carrying the ball across the opponent’s goal-line.  Rugby League has numerous similarities to American Football: it is played on a similar size pitch (100m x 70m) with a “set of 6 tackles” per possession (in place of  “4 downs”) in which a team has the opportunity to score a try. The scoring convention of Rugby League is: 4 points for a try and 2 points for the subsequent successful conversion; 2 points for a successful penalty kick; and 1 point for a successful field-goal. 

The most popular Rugby League competition in the world is the “National Rugby League (NRL)”, which consists of 16 teams across Australia and New Zealand. The NRL is the most viewed and attended Rugby League competition in the world. The 2018 NRL season was particularly noteworthy as it was the most competitive regular season in history of the league, and we believe in the history of sport. After 25 rounds of the regular season competition (where each team plays 24 games), the top 8 teams were separated by one win. Specifically, the top four teams (Roosters, Storm, Rabbitohs and Sharks) all finished with 16 wins. The next four teams (Panthers, Broncos, Warriors and Tigers) finished with 15 wins. The minor premiership –  which goes to the team that finishes first during the regular season was determined by point differential –  saw the Sydney Roosters win it for the fourth time in the past six years. To achieve this feat, the Roosters had to beat the Parramatta Eels by a margin of 27 points on the last day of the regular season, which they did with a 44-10 victory.  The result meant they pushed the Melbourne Storm out of top spot, with a point differential of just +8. Given the closeness of the competition, it would be useful to have many measurement tools to dissect how the league was won – which is what Rugby-Bot enables. 

\begin{figure}
\centering
    \includegraphics[width=0.8\linewidth,trim={0 0 0 0}, clip]{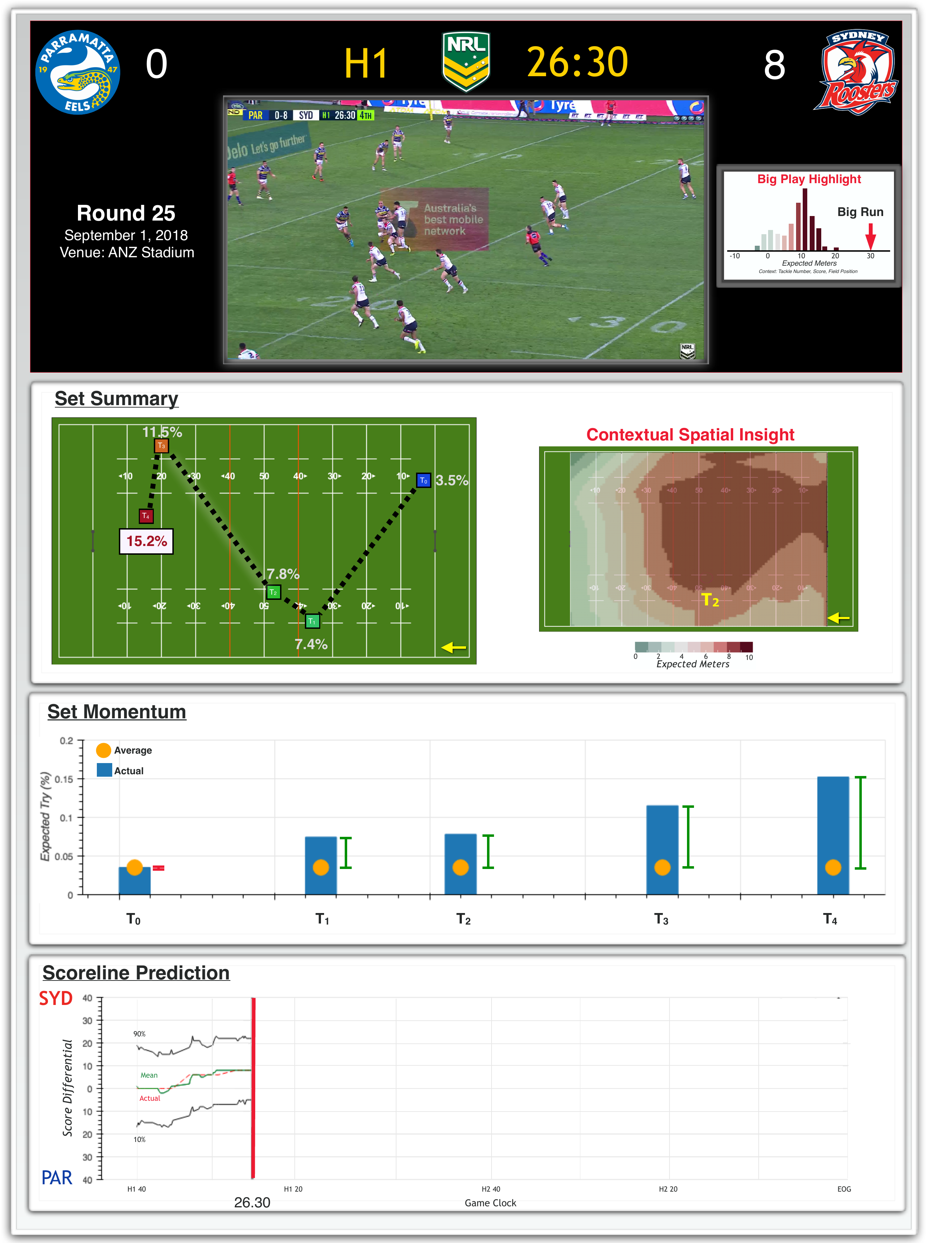}
\caption{An example of our Rugby-Bot tool used to analyze a Round 25 match from the 2018 NRL Season which consists of the following components: (Top) Shows the video and game information, as well as an indicator of a “highlight” using our anomaly detector. (Second Row) Shows the spatial information of each tackle in addition to the Expected TrySet Value for that set at each tackle.  Additionally, “Contextual Spatial Insights”, can be gleaned by looking at the expected meters gained for each location on the field under for a given context. (Third Row) The “Set Momentum” is monitored by tracking the current Expected Try Value at each tackle relative to average value. (Bottom Row) The final scoreline predictor tracks the current scoreline (red) and final prediction distribution (mean in green, 90-10\% confidence interval in black). }
\label{fig:1_rugbybot}
\end{figure}

An example of Rugby-Bot tool is depicted in Figure~\ref{fig:1_rugbybot}. In the example we show a complete “set” (i.e., Rugby League’s equivalent of a “drive” in American Football) from the Round of 25 of the 2018 NRL Season with the Roosters in possession against the Eels.  In the “Set Summary” (Second Row - left), Rugby-Bot tracks the likelihood the Roosters will score a try (exTrySet) during that set at the start of each tackle.  We see that they initially field the ball in terrible field position and only a 3.5\% likelihood of scoring.  However, a great return boosts their exTrySet to 7.4\% at the start of their first tackle.  A huge run during the second tackle raised their exTrySet further to 11.5\%, and good central field position on the fourth tackle sees it grow to 15.2\%.  Unfortunately for the Roosters, after a bad fifth tackle they elect to kick in an attempt to recover the ball and turn the ball over.  Rugby-Bot also allows us to analyze that decision making, which we will explore later.   We can dig further into the spatial context of the big run by exploring the expected meters gained as a function of field location in the “Contextual Spatial Insights” panel (Second Row – right).  This panel shows the expected meters as a function of not only the field position, but the full context: tackle number, team, scoreline, and opponent.  In the plot we see the advantaged gained by starting a tackle near the middle of the pitch.

An important concept in Rugby League is that of “Momentum” during a set of six tackles. This is done in Rugby-Bot by comparing the current exTrySet to the average for that context (field position, time, scoreline, team identities).  In the example shown in Figure~\ref{fig:1_rugbybot} (Third Row),  we see a boost in momentum both from the runback after the initial possession as well as the big run on tackle 2.  Rugby-Bot enables us to identify big runs not just in terms of the meters gained, but in how exceptional that run was given that context.  Next to the video player we see a “Big Play Highlight” indicator showing the distribution of meters gained expected under the conditions of tackle 2 and where the actual ran fell in this distribution, indicated by the red arrow. 

Finally, in the bottom row, we see the Scoreline Prediction tracker. In this plot, we not only track the current score differential (red dashed line), we predict the end-of-game score differential.  Note that because of our approach, which we will explore in the next section, our predictor is not just a final value, but a distribution, displayed here as the 90-10\% confidence interval (black lines) around the mean (green line).  We will dive into analyzing this plot more in Section~\ref{Analysis}.

\section{Multi-Task Learning using Fine-Grain Spatial Data}
Rugby-Bot is designed to analyze the state of the game across multiple time resolutions: play, set, and game.  That is, in any given situation, we would like to know how likely various outcomes are.  In particular, we predict:

\begin{itemize}
    \item Play Selection [$\play$]: likelihood a team will perform an offensive kick, defensive kick, run a normal offensive play on the next tackle
    \item Expected Meters (Tackle) [$\eMeters$]:  predicted meters gained/lost on the next tackle
    \item Expected Try (Tackle) [$\eTryTackle$]:  likelihood of scoring a try on the next tackle
    \item Expected Try (Set) [$\eTrySet$]:  likelihood of scoring a try at some point during that set
    \item Win Probability [$\winProb$]: likelihood of winning the game
    \item Scoreline Prediction [$\scoreline$]: predicted final score of each team
    \end{itemize}

Simply constructing a model “or specialist” for each task leaves much to be desired: it fails to share information and structure that is common across tasks and is an (unnecessary) hassle to maintain in a production environment.  Instead we strategically frame this problem as a model whose output can be thought of as a “state-vector” of the match.  

\subsection{National Rugby League (NRL) Dataset}
Rugby-Bot is built on data from the 2015-2018 National Rugby League seasons which includes over 750 games and more than 250,000 distinct tackles (i.e., plays).  For our analysis, the continuous event data was transformed into a segmented dataset with each data point representing a distinct tackle in the match.  At each tackle, we have information about the position of the ball, the subsequent play-by-play event sequence, the players and teams involved, the team in possession of the ball, as well as game-context (e.g. tackle number, score, time remaining). Data is split 80-train/20-test for modeling. 

\subsection{Modeling Approach}
\begin{figure}
\centering
    \includegraphics[height=0.3\linewidth,trim={0 0 0 0}, clip]{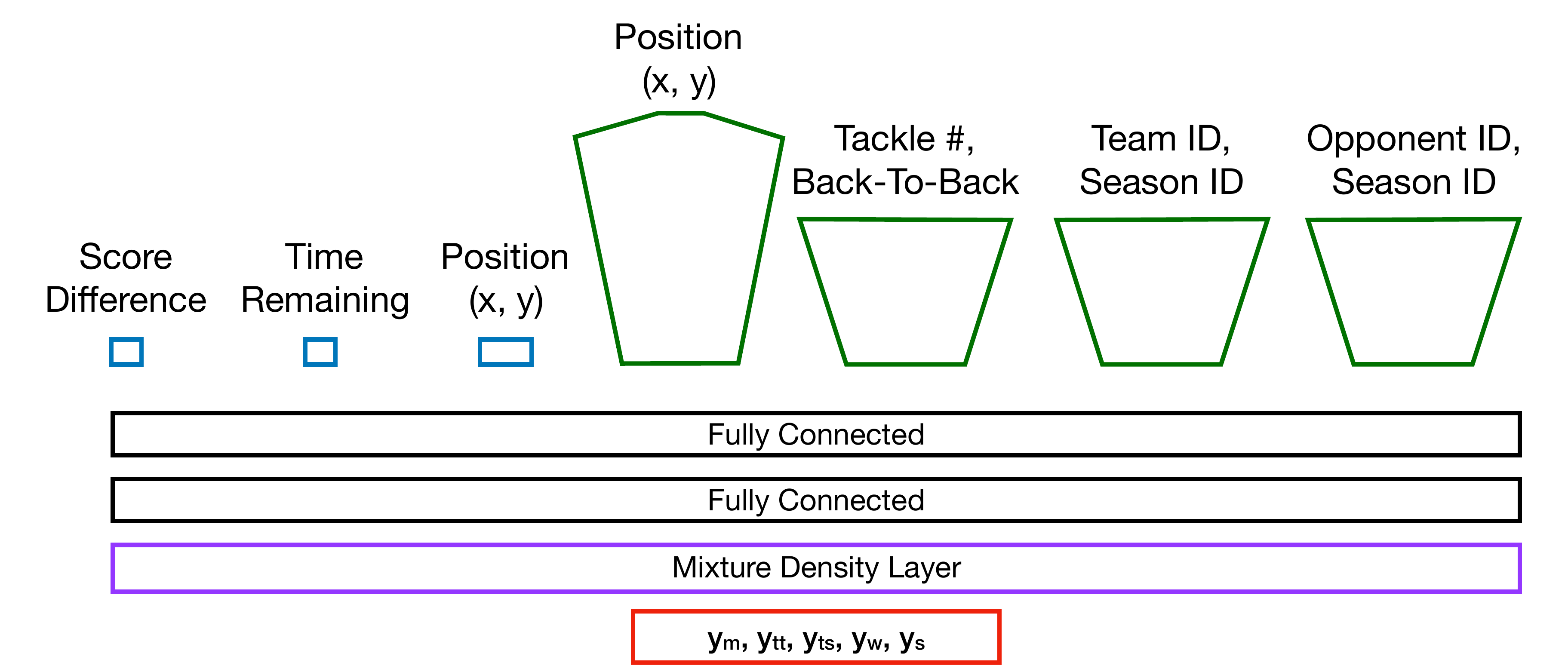}
\caption{Our base network architecture is a Mixture Density Network (MDN).  Categorical features as well as field position ($\x$, $\y$) are passed through embedding layers (green) to create a dense representation.  These are then concatenated with the continuous features (blue) and passed through two fully connected layers (black).  The mixture density layer consists of 5 mixture each with a prediction for $\mu$ and $\sigma$.  Expected meters ($\eMeters$), expected try tackle ($\eTryTackle$), expected try set ($\eTrySet$), win probability ($\winProb$), and final scoreline ($\scoreline$) are all predicted simultaneously.}
\label{fig:2_network}
\end{figure}

Our approach draws on three key approaches used throughout the machine learning and deep learning communities.  First, it draws on “wide and deep” models commonly used for recommendation tasks~\cite{cheng2016wide}.  Second, we cast this as a multi-task learning approach, predicting all outcomes from the same model, thereby sharing common features throughout.  Finally, we use a mixture density network which enables us to produce a distribution of outcomes, thereby capturing the uncertainty in the network.  The advantage to this is three-fold.  First, the Bayesian nature of MDNs afford us the ability to treat all losses in terms of a probabilistic likelihood and therefore enable us to do regression and classification tasks simultaneously.  Next, it is intrinsically multi-task which forces the model to learn globally relevant features about the game.  Finally, by modeling the outputs as a multi-dimensional distribution, it captures the relationships between the outputs as well as providing uncertainties about the predictions. The model architecture is shown in Figure~\ref{fig:2_network} and we will explore how each component contributes to the prediction task and the advantages that this architecture offers.

\subsubsection{Capturing Spatial and Contextual Information through Wide and Deep Models}
Our goal is to leverage both spatial and contextual information to make predictions about the outcome at the play, possession, and game-level.  Some of the challenges in modeling this data include: 

\begin{itemize}
    \item Raw spatial data is low-dimensional and densely represented
    \item Spatial features are inherently non-linear
    \item Contextual features (e.g. teams, tackle) are discrete and may be sparse
    \item The relationship between the spatial and contextual features is non-linear
\end{itemize}

To address these points, we draw on inspiration from the wide and deep models of~\cite{cheng2016wide}.  Following the approaches there, we pass each categorical feature through its own deep portion of the network to construct higher-level, dense features from the originally sparse inputs, creating an embedding vector.  These embeddings are then concatenated with the remaining dense features (such as time and position) before passing through 2 final dense layers to capture their non-linear relationships.  

A uniqueness to our approach is in how we represented the interactions of certain categorical (i.e. contextual) features such as season and team/opponent identity.  As is common practice, we represented each as a one-hot vector.  As opposed to creating an embedding for season, team, and opponent independently, however, we instead concatenated the one-hot vector of season with the one-hot vector of team (and similarly for opponent).  This enables us to more directly share identity information across seasons, capturing the notion that teams maintain a level of similarity across years while also picking up on league-wide trends over different seasons.  We used a similar approach to capture the tackle number (1-6) and back-to-back sets (represented as a binary flag).  

The other key innovation in our approach is the manner in which high-level features are extracted from the field position; this addresses the first point above.  The field position is a dense, two-dimensional ${\x,\y}$ input.  In a traditional wide and deep model, this value would simply be concatenated at the embedding layer.  However, the shallowness of the wide portion of these models prevents extraction of higher-level spatial features.  To address this, we treat field position as both a sparse and dense feature.  

For the sparse portion, we pass the two-dimensional position through several layers just like the categorical features.  Note that we initially boost from the original two-dimensions up to 50 at the first hidden layer.  This gives the network sufficient dimensionality to “mix” the inputs and create the desired higher-level features.  This is a key insight.  Traditionally, the low dimensionality of the spatial input has been addressed by fully discretizing the field position, that is, by treating the field as a grid and representing the position as a 1-hot vector indicating the occupancy in that bin~\cite{DBLP:conf/icdm/YueLCBM14, miller2014factorized}.  This has several limitations.  First, the discrete representation is inherently lossy as positions are effectively rounded during the binning process.  Second, the resulting dimensionality may be massive (e.g. a 70m x 100m field broken into $\frac{1}{3}$ m $\times$ $\frac{1}{3}$m bins results in a 63,000 dimensional input).  While we would like to increase our dimensionality above 2, expanding to the thousands or tens-of-thousands is unnecessarily high and results in an extremely sparse representation, requiring more neurons and more data.  Our approach avoids these pitfalls by working directly on the raw spatial input, moderately increasing the dimensionality in the first hidden layer, and then enabling the network to extract important spatial relationships directly from the data.

Finally, because of the importance of the field position in all of these prediction tasks, we also include the position again as a dense feature.  This enables us to capture both low-level (i.e. raw), as well as high-level (i.e. embedded) spatial information.  

\subsubsection{Mixture Density Network for simplified Multi-Task Learning}
Multi-task learning has been commonplace within the machine learning because of its ability to exploit common structures and characteristics across tasks~\cite{caruana1997multitask}.  Often a difficult with multi-task learning comes when one would like to perform both regression (which would use a loss such as RMSE) and classification (which would use a loss such as cross entropy) simultaneously.  For example, in predicting both win probability and final score-line, how much should a 1\% likelihood in winning be weighted relative to a 1pt change in the score?  Even when all tasks are of the same type (classification or regression), how each loss is scaled and/or weighted can be somewhat arbitrary: how should 1m error in the predicted meters gained be waited against 1pt error in the final scoreline?  

Mixture density networks (MDN) are a class of models which combine neural networks and mixture density models~\cite{bishop_MDN}.  As such it is naturally Bayesian and allows us to model the conditional probability distribution $p(\parameters|\inputVec)$ where $\parameters$ are the parameters which describe the distributions generating our game-state vector $[\play, \eMeters, \eTryTackle, \eTrySet, \winProb, \scoreline]$  and $\inputVec$ is our input feature vector.  Therefore, we have a single loss for our prediction: the negative log-likelihood that a game-state is observed given the current game-state.  This dramatically simplifies our loss function.  Additionally, the loss over the full distribution constrains the model to learn the relationships between the outputs, for example, that it would be unlikely to observe a high value for meters gained but very low value for an expected try.

The “mixture” part of the MDN allows us to capture the multi-modality of the underlying distribution.  This produces various “modes” of the game, such as lopsided contests or very close scenarios.  MDNs also give us insight into the uncertainty of our models “for free”.  This enables us to generate a distribution of outcomes - as opposed to just a mean - for any given prediction. 

Cross entropy is used as the loss for training.  Our test loss is .051.  Note that this is a single loss for across the entire state-vector: even if several components in a prediction are very good, the loss may be quite high if the state is inconsistent (e.g. a positive final scoreline prediction, but a low win probability are inconsistent and therefore would have a high loss).

\subsubsection{Modeling decision-making on the “last tackle”}
Similar to the decision-making opportunities on 4th-down in American Football (punt, kick, go-for-it), in Rugby League teams must decide whether to kick defensive (equivalent to a punt), kick offensively (in an attempt to regain the ball for back-to-back sets), or “go-for-it” on the last tackle.  For this task we use a simple logistic regressor.  We elect to model this decision separately because it only pertains to the last tackle and is therefore “silent” on most downs unlike the other tasks which are active every single tackle.  The test loss for this model is .64.

\section{2018 NRL Season Analysis: DVOA for Rugby League}
Football Outsider’s DVOA (Defense-adjusted Value Over Average) has become arguably the most popular advanced statistic in the NFL. It aims to correctly value each play and its impact on scoring and winning.  DVOA factors in the context such as opponent, time, and score to compare performance to a baseline for a situation~\cite{dvoa}.  It can be broken down and applied in several ways, but the most popular application is to measure team strength. 

Rugby-Bot can be used to create to a DVOA for Rugby League.  Like American Football, Rugby League has segmented plays where the fight for field position is key and not all yards/meters are equal even though they appear so in the box score. The Expected Try (set) prediction generates the predicted chance of scoring during the set for every play. The prediction, like football’s DVOA, considers the score, time, opponent, and location.  Taking the difference between this expected value and the actual value we can see which teams outperform their expectation.  To create a more interpretable ‘DVOA’ the values are scaled with a mean of 0.  A positive offensive DVOA and negative defensive DVOA correspond to strong offenses and defenses, respectively.

We can use our Rugby League DVOA to analyze the 2018 NRL Season as seen in Table~\ref{tab:1_leagueSummary}. 2018 was an extraordinarily close season in which the top 4 teams all had the same number of points and wins.  The top two teams had similar meters for and against, however, the context under which those meters were gained and conceded strongly favored the Roosters as indicated by their DVOA differential.  The champion Roosters had the top DVOA per play mostly due to their strong defense, surpassing the next strongest defensive team (the Storm) by more than 1.5 defensive DVOA pts per play.

\begin{table}
 \label{tab:1_leagueSummary}
 \caption{The points, wins, losses, meters for and against, meter differential (for-against), offensive and defensive DVOA, and differential DVOA (offensive-defensive) is shown for each team during the 2018 NRL season.   A large positive number for Offensive DVOA and a large negative number for Defensive DVOA indicate a strong offense and defense, respectively.}
  \centering
  %\begin{tabular}{lll}
  \resizebox{\textwidth}{!}{\begin{tabular}{l l l l l l l l l l l}
    \begin{filecontents*}{./figures/tab1.csv}
Rank,Team,Pts,Wins,Losses,Meters For,Meters Against,Meters Diff,DVOA Offense,DVOA Defense,DVOA Diff
1,Roosters,34,16,8,542,361,181,1.62,-5.75,7.37
2,Storm,34,16,8,536,363,173,0.57,-4.17,4.74
3,Rabbitohs,34,16,8,582,437,145,2.37,-0.97,3.35
4,Sharks,34,16,8,519,423,96,1.72,-1.84,3.56
5,Panthers,32,15,9,517,461,56,0.39,-1.67,2.05
6,Broncos,32,15,9,556,500,56,0.99,0.29,0.7
7,Dragons,32,15,9,519,472,47,3.62,1.92,1.7
8,Warriors,32,15,9,472,447,25,1.96,0.91,2.87
9,West Tigers,26,12,12,377,460,-83,-2.77,1.41,-1.35
10,Raiders,22,10,14,563,540,23,3.62,1.92,1.7
11,Knights,20,9,15,414,607,-193,-0.28,3.6,-3.88
12,Bulldogs,18,8,16,428,474,-46,-3.57,0.29,-3.86
13,Cowboys,18,8,16,449,521,-72,-2.58,0.95,-3.54
14,Titans,18,8,16,472,582,-110,-0.85,3.74,-4.59
15,Sea Eagles,16,7,17,500,622,-122,1.5,4.42,-2.88
16,Eels,14,6,18,374,550,-176,-3.96,2.15,-6.11
\end{filecontents*}
\csvautotabular{./figures/tab1.csv}
  \end{tabular}}
\end{table}

\begin{figure}
\centering
    \includegraphics[width=\linewidth,trim={0 2.2cm 0 0}, clip]{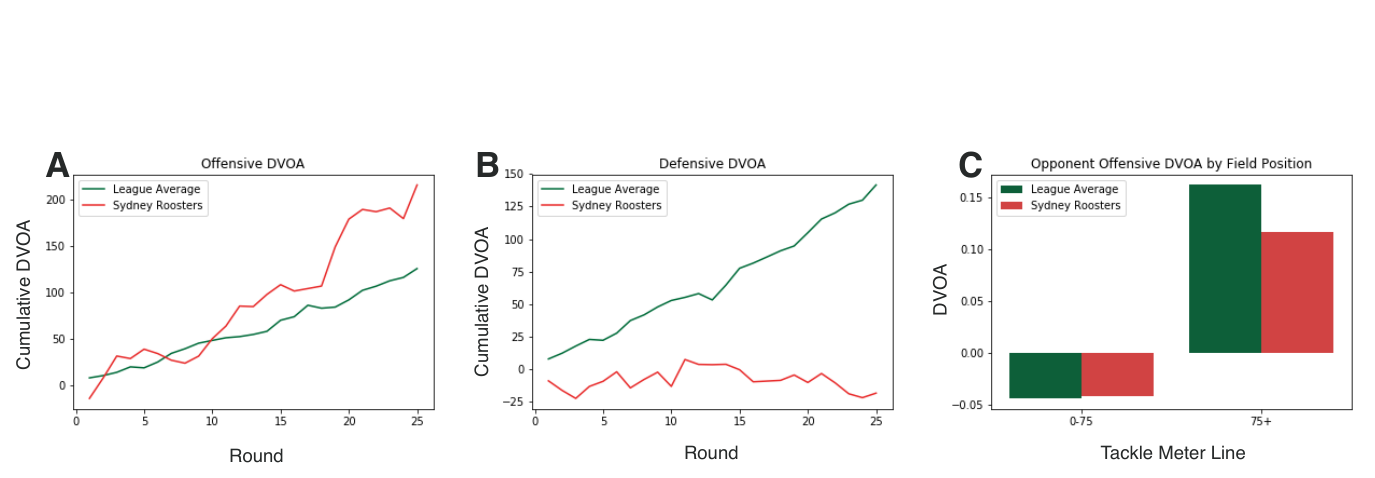}
\caption{(A) The cumulative Offensive DOVA for the Roosters (red) and the League-average (green); the Roosters got off to a slow start offensively then began to pull away from the field.  (B) The cumulated Defensive DVOA for the Roosters (red) and League-average (green); the Roosters dominated defensively (indicated by a negative value for defense) through-out the season.  (C) Comparison of the Roosters to the League when their opponents were on offense and in normal field position versus the near the goal (within the 75m line).}
\label{fig:3_dvoa}
\end{figure}

Next we examine the Roosters form over the course of the season.  Figures~\ref{fig:3_dvoa}A and \ref{fig:3_dvoa}B break down the offensive and defensive strength over the season.  In Figure~\ref{fig:3_dvoa}A we see that initially the Roosters got off to a slow start, largely on-par with the league average (they started the season with a 4-4 record).  However, by mid-season, they began to find their form. The offensive DVOA allows us to track how the Roosters went from average to great in the 2nd half of the season.  Defensively, the Roosters were dominant all season long as seen in Figure~\ref{fig:3_dvoa}B.  They maintained their strong form even as the league worsened in Defensive DOVA.

With our individual play valuation we can dive deeper in the Roosters success and reveal where their defense outperformed their opponents.  Figure~\ref{fig:3_dvoa}C shows that the Roosters defense was average during plays occurring in the first $\frac{3}{4}$ of the field.  (Note that both the Roosters and the league average have a negative DVOA in these situations as they rarely result in scoring.)   However, in critical moments, in the final 25m of the pitch, they were incredibly strong in the last 25 yards with their Defensive DVOA being maintained at 0.12 while the league-average shot to over 0.15 in key situations.

\section{Rugby League Game and Play Analysis}
\label{Analysis}
\subsection{Distributed Scoreline Prediction: Intelligent Game Analysis}
We return now to the critical Round of 25 matchup between the Roosters and the Eels which was shown earlier.  Recall that to clinch the title, the Roosters needed to win this match by 27 points.  In Figure~\ref{fig:1_rugbybot}, we viewed final scoreline prediction only a few minutes into the match when the Roosters lead by only 8 pts.  Is the title within reach?  To answer this question we need to be able to predict not only the final score differential, but the uncertainty around this measurement.  Rugby-Bot enables us to do this.  Similar to the approaches used in ~\cite{dvoa} our MDN allows us to predict a range of likely final scorelines at each point in the match.  

\begin{figure}
\centering
    \includegraphics[width=\linewidth,trim={0 0 0 0}, clip]{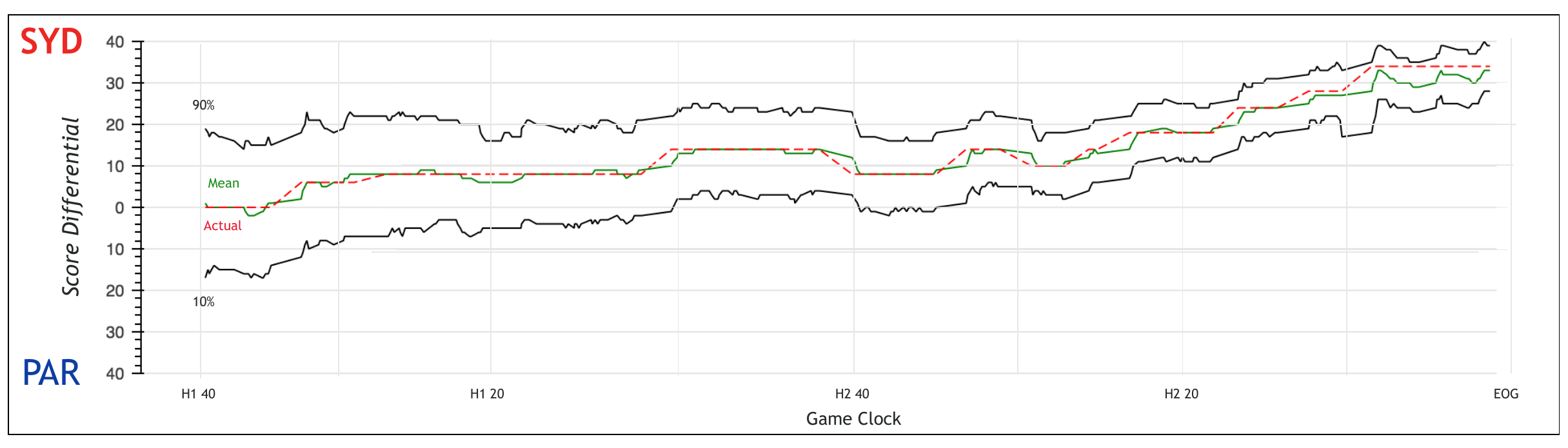}
\caption{(A) The predictor of the final scoreline for the Round of 25 matchup between the Roosters and Eels.  The actual score is shown in red, the mean of the predicted distribution is shown in green, and the 90-10\% confidence interval is shown in black.}
\label{fig:4_scoreline}
\end{figure}

Now we plot the full scoreline predictor over the course of the match in Figure~\ref{fig:4_scoreline}.  The actual score is shown in red and the predicted mean score in green with the 90-10\% confidence interval shown in black around it.  Note that although we are showing the mean here, the mean need not be the most likely value; as our model is a mixture of gaussians it is inherently multimodal and the most likely outcome could occur at a mode away from the mean.   At any point in the game we can view the full shape of the predicted distribution, but here the mean and 90-10 bounds are presented for visual simplicity.  Also note that because it is a mixture model, the distribution complex and not necessarily symmetric: the values of the mean, 90-bound, and 10-bound trend together, but are not identical.  

There are several interesting observations we can draw from this analysis.  First, with 10 minutes remaining in the first half, the game appears to be largely over: the lower 10\% bound has moved about a score differential of 0, making a comeback unlikely.  It dips back down ever so slightly after the start of the second half and stays here for the next few minutes until by H2 33:32, the 90-10 interval never includes 0 for the remainder of the game.  At this point we can say the same is effectively over.  This is interesting in recreating the story of the game, but is also useful for analyses such as quantifying “garbage time” or understanding the risk impact of substitutions to rest key players late in a game.

Although the game is “over” shortly into the second half, the fate of the season is still uncertain.  The Roosters must win by 27 and at H2 33:32, they are up only by 14.  Furthermore, even the 90-10 confidence interval does not include a 27 point victory for the Roosters.  This does not occur until H2 18:13 when the Roosters go up by 22.  Now the title is within reach.  

Note that as the game progresses, the width of the confidence interval shrinks.  Early in the game the outcome is very uncertain, but as the game progresses, the range of possible outcomes shrinks until it is identically known once the game is over.  Even in the remaining few minutes, there is still some uncertainty as to the final score, as there is always a possibility for a last-minute score or turnover.  However, the model has learned that at most the score will fluctuate by is the value of a single try and conversion.  

\subsection{Valuing and Comparing Individual Plays Across Different Contexts}
\begin{figure}
\centering
    \includegraphics[width=\linewidth,trim={0 0 0 0}, clip]{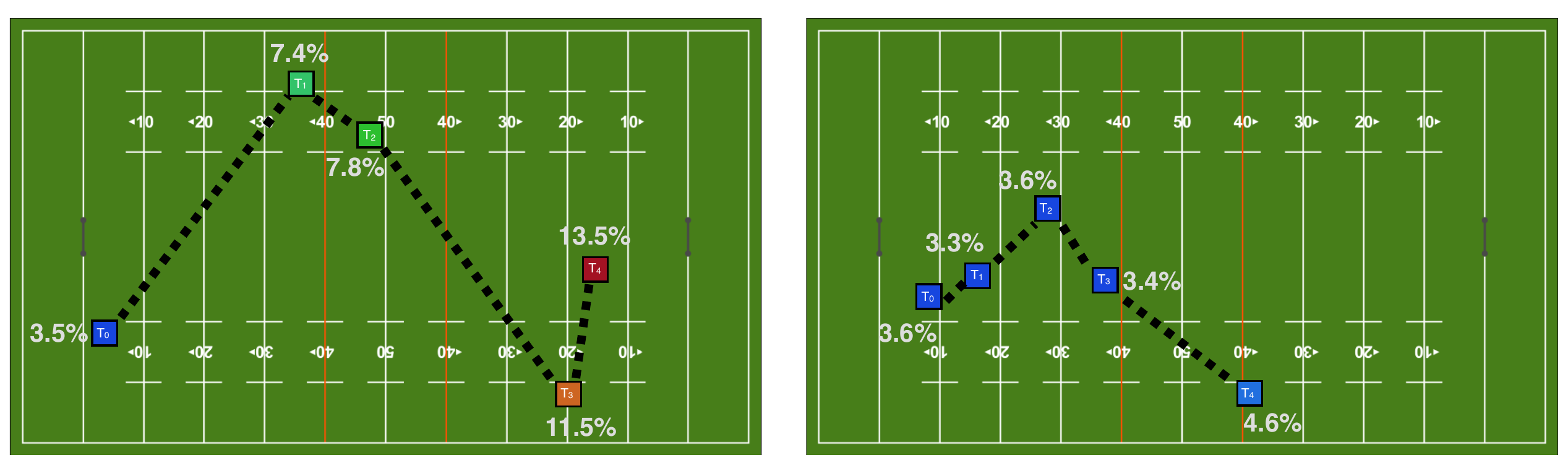}
\caption{The likelihood of scoring a try at each tackle (exTrySet) is shown for two different sets.  Plays have been normalized so the offensive team is always going left to right.}
\label{fig:5_playcompare}
\end{figure}

Rugby-Bot allows us to determine the value of each play in a game. To demonstrate this, we can compare two plays from the Round 25 Eels v Roosters Game. It is obvious that a big run will increase a team’s chance of scoring, but not all meters are created equal.  Figure~\ref{fig:5_playcompare} shows two different sets in the match; on the left is the same set in Figure~\ref{fig:1_rugbybot}, and on the right is another set occurring near the end of the first half. 

We can evaluate tackle 3 from the example shown on the left, and tackle 4 in example shown on the right in Figure~\ref{fig:5_playcompare}.  Both runs were over 22 meters and over 95th percentile of predicted expected meters.  In example shown on the left, the run is made closer to the goal line and earlier in the set. This causes the exTrySet value to go up 3.8\% while the run in example shown on the right is late in the set and far from the goal line so it only increases the exTrySet 1.2\%.  This quantifies the conclusion that an observer would create that the meters gain in the example of the right were not as valuable as the chances of scoring are still miniscule. Taking these deltas between plays allows us to pinpoint the plays in the match that are impactful outside of just yards or scoring.

\begin{figure}
\centering
    \includegraphics[width=0.5\linewidth,trim={0 1.0cm 0 0}, clip]{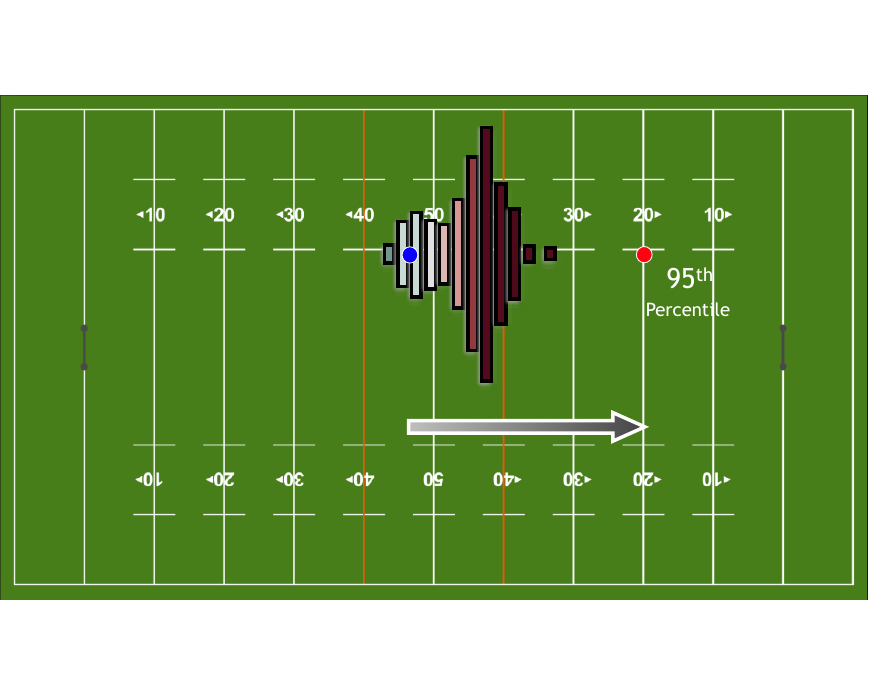}
\caption{The blue dot corresponds to the starting position of the play and the red dot is where the play ended up. Using our distribution analysis, we can see that this play was in the top 5\% of plays (or in the 95th percentile).}
\label{fig:6_lasttackle}
\end{figure}

To further highlight how good this play on the left of Figure~\ref{fig:5_playcompare} was – we can use a distribution of predictions to easily identify big plays and analyze individual play performance.  Figure~\ref{fig:6_lasttackle} visualizes the 22 meter run through a different lens. The blue dot corresponds to the starting position of the play and the red dot is where the play ended up. We can see based on the context $(\x,\y)$ starting position, tackle number and game context), this play was in the top 5\% of plays (or in the 95th percentile). Effectively, we can value each play and determine whether it was within the “top-plays” in any situation.

\subsection{Analyzing Decision Making on the Last Tackle}
The first five tackles in a set in Rugby League are largely uniform in their objective: score a try or gain the most meters possible towards the tryline.  On the last tackle, however, teams may elect to kick defensively (similar to a punt in American Football), kick offensively (in an attempt to score or regain the ball for back-to-back sets), or run a standard play in an attempt to score as they have to turn the possession of the ball over to the opposition at end of the tackle.  Rugby-Bot enables us to predict what a given team is likely to do, analyze the outcome of that decision, and assess the impact of that decision on the game relative to another choice having been made.

\begin{table}
 \caption{Within the opponent’s 20 meter line, teams consistently make conservative decisions, electing to kick when running produces more than twice the expected points.}
  \centering
  \begin{tabular}{lll}
    \toprule
    \cmidrule(r){1-3}
    Play Decision     & Frequency     & Expected Points \\
    \midrule
    Run     & 32\%  & 1.15     \\
    Kick     & 68\% & 0.53      \\

    \bottomrule
  \end{tabular}
  \label{tab:2_decision}
\end{table}

Given the last tackle is often the most exciting and variable play – it enables us to investigate some interesting behavioral questions. One such question is “do teams make the best decision on the last tackle, or are they naturally conservative in their play calling?”  To answer this requires us to analyze both the position as well as context surrounding the sixth tackle.  Using our predictions of play call (run, defensive kick, offense kick) and our play values, we can value each decision on each part of the field.  In Table~\ref{tab:2_decision}, we see that in the “Red Zone” attempting to run the ball for a Try is more valuable than kicking.  Despite this fact, teams tend to be conservative and attempt to run it on only 32\% of plays.

Table~\ref{tab:3_lastTacklePerformance} displays how the most aggressive team in the 2018 season, the Canberra Raiders, also have the most points over expected. Two of the best teams, the Roosters and Storm, do not run the ball often on final tackle. This could be because, as more talented teams, they can afford to be risk averse.

\begin{table}
 \label{tab:3_lastTacklePerformance}
 \caption{2018 Season performance on last tackle.  Each team has roughly the same Expected Points on their final tackle, but the actual points obtained varies dramatically.}
  \centering
  \resizebox{\textwidth}{!}{\begin{tabular}{lcccccc}
  \toprule
    Team	& Expected Points	&Actual Points	& Over Expected	& Run$(\%)$	& Run$(\%)$ Rank & Over Expected Rank \\
  \midrule
    Canberra Raiders                & 0.56	& 1.04	& 0.47	& 41.67\%	& 1	& 1 \\
    Brisbane Broncos                & 0.54	& 0.97	& 0.43	& 37.18\%	& 6	& 2 \\
    Penrith Panthers                & 0.56	& 0.78	& 0.22	& 38.39\%	& 5	& 3 \\
    Newcastle Knights               & 0.56	& 0.73	& 0.18	& 39.80\%	& 2	& 4 \\
    Cronulla-Sutherland Sharks	    & 0.52	& 0.68	& 0.16	& 30.58\% & 9	& 5 \\
    Manly-Warringah Sea Eagles	    & 0.55	& 0.66	& 0.11	& 25.89\%	& 10	& 6 \\
    North Queensland Cowboys	    & 0.56	& 0.65	& 0.09	& 19.58\%	& 15	& 7 \\
    South Sydney Rabbitohs	        & 0.55	& 0.6	& 0.04	& 24.47\%	& 13	& 8 \\
    Sydney Roosters	                & 0.58	& 0.57	& -0.01	& 16.16\%	& 16	& 9 \\
    Gold Coast Titans               & 0.54	& 0.5	& -0.04	& 38.46\%	& 4	& 10 \\
    Canterbury Bulldogs	            & 0.55	& 0.51	& -0.04	& 33.79\%	& 8	& 11 \\
    St. George Illawarra Dragons	& 0.55	& 0.51	& -0.05	& 39.39\%	& 3	& 12 \\
    Melbourne Storm	                & 0.53	& 0.48	& -0.06	& 25.24\%	& 11	& 13 \\
    New Zealand Warriors	        & 0.53	& 0.46	& -0.07	& 20.56\%	& 14	& 14 \\
    Wests Tigers	                & 0.56	& 0.42	& -0.13	& 37.17\%	& 7	& 15 \\
    Parramatta Eels	                & 0.55	& 0.38	& -0.17	& 25.00\%	& 12	& 16 \\
    \bottomrule
  \end{tabular}}
\end{table}

Returning to our example set from Round 25 (Figure~\ref{fig:1_rugbybot}), we saw the Roosters in outstanding field position before the final tackle, with an exTrySet value well above average.  However, instead of running the ball, which they had done with great success during that set, they elected to kick to the far touch-line, consistent with their style of running far less than any other team in the league on the final tackle.  The ball was mishandled, lost out of bounds, and possession was lost to the Eels.  The Eels mounted an extended drive to the shadow of the Roosters’ goal, only to fail to convert.  The Roosters’ defense again came to their rescue.   Overall, the analysis shown in Table~\ref{fig:3_dvoa}is effectively the “Red Zone efficiency”, which teams can utilize to see how effective they are in the most important area of the field.

\section{Summary}
In this paper, we presented a multi-task learning method of generating multiple predictions for analysis via a single prediction source. To enable this approach, we utilized a fine-grain representation using fine-grain spatial data using a wide-and-deep learning approach. Additionally, our approach can predict distributions rather than single point values. We highlighted the utility of our approach on the sport of Rugby League and call our prediction engine “Rugby-Bot”. Utilizing our slue of prediction tools, we showed how Rugby-Bot could be used to discover how the Sydney Roosters won the 2018 NRL minor premiership (the closest in the history of the sport). Additionally, we showed how it can be used to value each individual play, accurate predict the outcome of the match as well as investigate play behavior in Rugby League – specifically behavior on the last tackle.

\bibliographystyle{unsrt}  
\bibliography{references}  %%% Remove comment to use the external .bib file (using bibtex).

\end{document}